\documentclass[a4paper,conference]{IEEEtran}

\usepackage[pdftex]{graphicx}
\usepackage{subfig}
\usepackage{array}
\usepackage{multirow}
\usepackage{color}
\usepackage{pifont}
\usepackage[linesnumbered,ruled,vlined]{algorithm2e}
\usepackage{amssymb}
\usepackage{hyperref}
\hypersetup{colorlinks=true,urlcolor=blue}

\SetCommentSty{mycommfont}

\makeatletter
\algocf@newcommand{InputXX}[1]{%
  \sbox\algocf@inputbox{\hbox{\KwSty{Output}\algocf@typo: }}%
  \ifthenelse{\boolean{algocf@inoutnumbered}}{\relax}{\everypar={\relax}}%
  {\let\\\algocf@newinput\hspace{\wd\algocf@inputbox}\hangindent=\wd\algocf@inputbox\hangafter=\wd\algocf@inputbox#1\par}%
  \algocf@linesnumbered
}
\makeatother

\usepackage{xspace}
\makeatletter
\DeclareRobustCommand\onedot{\futurelet\@let@token\@onedot}
\def\@onedot{\ifx\@let@token.\else.\null\fi\xspace}

\def\ie{\emph{i.e}\onedot}

\def\etal{\emph{et al}\onedot}
\makeatother

\graphicspath{{imgs/}}

\begin{document}

\title{Crack Detection as a Weakly-Supervised Problem: Towards Achieving Less Annotation-Intensive Crack Detectors}

\author{\IEEEauthorblockN{Yuki Inoue, Hiroto Nagayoshi}
\IEEEauthorblockA{Hitachi Ltd.\\
Tokyo, Japan\\
Email: \{yuki.inoue.wh, hiroto.nagayoshi.wy\}@hitachi.com}
}



\maketitle

\begin{abstract}
    Automatic crack detection is a critical task that has the potential to drastically reduce labor-intensive building and road inspections currently being done manually. Recent studies in this field have significantly improved the detection accuracy. However, the methods often heavily rely on costly annotation processes. In addition, to handle a wide variety of target domains, new batches of annotations are usually required for each new environment. This makes the data annotation cost a significant bottleneck when deploying crack detection systems in real life. To resolve this issue, we formulate the crack detection problem as a weakly-supervised problem and propose a two-branched framework. By combining predictions of a supervised model trained on low quality annotations with predictions based on pixel brightness, our framework is less affected by the annotation quality. Experimental results show that the proposed framework retains high detection accuracy even when provided with low quality annotations. Implementation of the proposed framework is publicly available at \url{https://github.com/hitachi-rd-cv/weakly-sup-crackdet}.
\end{abstract}

%
\IEEEpeerreviewmaketitle

\section{Introduction} \label{intro}

The importance of crack detection is ever more highlighted by the recent rapid growth in the number of old roads and buildings.
Naturally, most of the studies focus on improving the detection accuracy \cite{fan_automatic_2018, li2018automatic, shi, zhang_temple, yang2019feature, liu2019deepcrack}, as distinguishing crack patterns from complex surface textures is a challenging task. 
As a result, recent works show high detection rates for many datasets, at least for cases in which train and test domains match. Therefore, we believe more focus should be placed on other aspects of crack detection.

When deploying crack detection systems in real life, one of the biggest bottlenecks is the cost required to annotate the collected data. Since crack detection problem is formulated as a semantic segmentation problem, pixel-level supervision, which is one of the most cost-intensive annotations to obtain, is required.
In addition, training a robust crack detector that can be used across all situations is unfortunately difficult, as the definition of a crack often varies across sites, and also crack patterns vary greatly across different materials. For example, a straight, black line on a wall defined as a crack at Site A will not be considered as a crack at Site B if the walls at Site B are made of stone tiles and inherently contain black line patterns.
Therefore, new sets of annotations are required per site to train site-specific detectors in order to maintain high detection accuracy across all sites, further increasing the annotation burden.

Recognizing this issue, we have decided to focus on reducing the annotation cost while maintaining accuracy. To this end, we summarize our contributions as follows:

\begin{itemize}
  \item We formulate the crack detection problem as a weakly-supervised task, and provide a suite of synthetic annotations as well as two manual annotations for this new task.
  \item We propose a framework that could act as a strong baseline for the task. Our proposal introduces a simple component that can be added to any semantic segmentation networks to extend their performance under weakly-supervised settings.
  \item We show that the proposed framework is effective under weakly-supervised settings, through thorough experiments on four network architectures and three datasets.
\end{itemize}

\section{Related Works}

\subsection{Crack Detection Approaches} \label{subsec_detection_approaches}

In general, there are two crack detection approaches:

\textit{Rule-Based Approaches}: These are approaches that use human-defined rules to make detection decisions. The most natural method is to use some form of edge detection \cite{morph_model, canny_model}. Others approach the problem by converting input images to other domains. For example, Subirats \etal applied the wavelet transformation on input images and isolated the crack regions in the frequency domain \cite{wavelet_model}.

One of the notable subset of this approach is the \textit{minimal path algorithm} (MPA), which models an image as a graph with crack regions being the edges, and keypoints, identified by detectors such as Histogram of Oriented Gradients (HOG), as vertices \cite{mpa2, mpa1}. The crack detection problem is then reduced to finding the most optimal edges between the vertices. For example, multi-scale fusion crack detection (MCFD) is an improvement on MPA, and it uses multiple scales of an input image to refine the result \cite{li2018automatic}.
Unfortunately, most of the rule-based algorithms are not robust, requiring manual recalibration or overhaul to fit with the new environments, or are too computationally intensive to be used in a reasonable time frame.

\textit{Learning-Based Approaches}: These approaches use data samples to tune the model's inner parameters. Deep learning based methods are one of the most actively studied methods for this category. Examples include the work of Zhang \etal, in which a shallow CNN was used for prediction \cite{zhang_road_2016}, and work by Fan \etal, in which the idea of structured prediction is introduced to force the model to learn the relationship between neighboring pixels \cite{fan_automatic_2018}. Inoue \etal proposed the Multiple Instance Learning architecture (MIL) which resembles test time augmentation (except it is also applied during training time as well) to increase the rotational robustness of the network \cite{inoue}.
Other works use deeper CNNs with encoder-decoder architecture with skip connections, to better fuse information from multiple scales \cite{yang2019feature, liu2019deepcrack}.

\subsection{Weakly, Semi, and Unsupervised Anomaly Detection} \label{subsec_semi_un_detection}

As mentioned in Sec. \ref{intro}, one of the biggest bottlenecks when deploying a crack detection system is the annotation cost. In this section, we consider three approaches generally taken to reduce the annotation cost.

The first approach is to reduce the annotation quality and formulate the problem as a weakly-supervised problem. 
By reducing the quality of the annotation, it becomes possible to annotate samples more quickly, reducing the annotation time.

For example, Fan \etal \cite{fan2019road} trained their model with image-level annotations, in which only the existence of a crack within an image is annotated, therefore lacking the precise location information.
One major disadvantage of their proposal is that the supervised model can only isolate the crack regions up to rectangular patches. As these selected patches mostly contain non-crack region because cracks are thin, the post-processing rule-based method needs to be very accurate, which is often difficult as mentioned in Sec. \ref{subsec_detection_approaches}.

The second approach is to annotate some samples, and formulate the problem as a semi-supervised problem. One typical method for semi-supervised learning is self-training, where the thresholded prediction values are used as the ground truth for unlabeled samples. Examples of other approaches include using the high-entropy nature of anomaly samples \cite{ruff2019deep}, and using agreement among the ensembled models \cite{LaineA17}.

The final approach completely eliminates the need for annotation, and formulates the problem as an unsupervised problem.
Examples of this approach include the rule-based methods mentioned in Sec. \ref{subsec_detection_approaches}. More recently, machine-learning based methods such as ones based on auto-encoders \cite{mujeeb2019one, kang2018deep} and Generative Adversarial Networks \cite{yu2020unsupervised} have appeared. In these methods, models learn to reconstruct undefected images. Crack detection is achieved through the model's inability to accurately reconstruct defected images.

\section{Method}

\begin{figure}[!tb]
	\centering
	\includegraphics[width=1\linewidth]{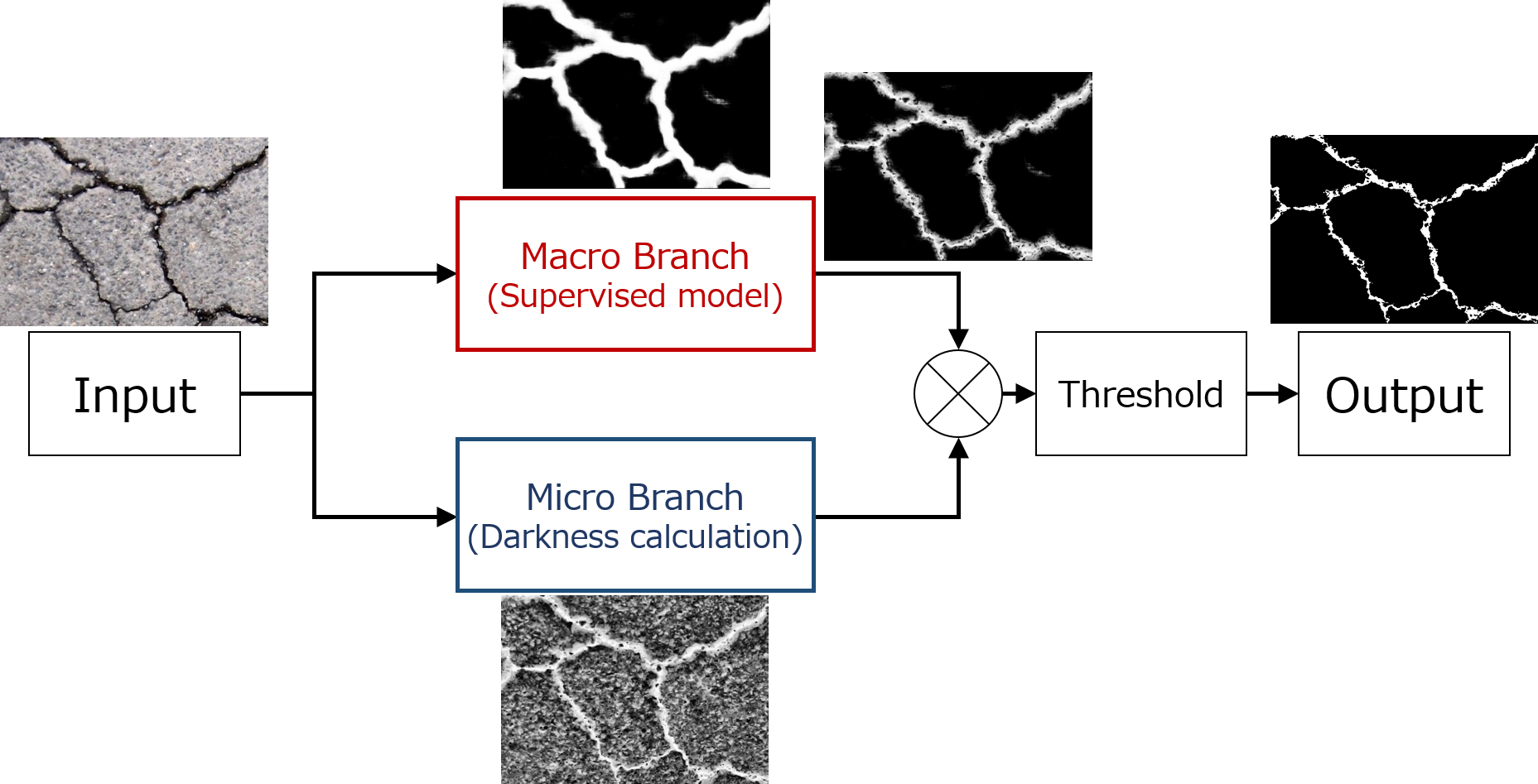} 
	\caption{Overview of the proposed framework with an example input.}
	\label{overall_framework}
	\vspace{-0.4cm}
\end{figure}

\subsection{Annotation Reduction Strategy} \label{subsec_choice}

As mentioned in Sec. \ref{subsec_semi_un_detection}, there are three approaches to reduce the annotation cost, namely weakly, semi, and unsupervised approaches. 
In this paper, we will formulate the crack detection problem as a weakly-supervised problem, for the following reasons.

First, unsupervised approach was not selected because although the performance of unsupervised methods have improved in the past years, they are still inaccurate when compared against the supervised counterparts. For example, Yu \etal \cite{yu2020unsupervised}, who claims to be the state of the art unsupervised crack detector, underperforms by 5.5\% in F1-score against a supervised model (the result shown in the paper is a little misleading, as the result of when the train and test datasets match is shown on the same table as the result of when the train and test dataset do not match).
This lack of performance can lead to cracks being overlooked, which in turn may lead to serious accidents. So if the small cost of creating rough annotations can lead to significantly better performance, it is strongly preferred over unsupervised approach.
In addition, strictly speaking, unsupervised models are not trained to detect cracks, but rather trained to detect \textit{any unforeseen differences} from the images they are trained with.
Therefore, they are more likely to misjudge non-crack changes as cracks than supervised models would. This is not desirable, because many of the changes applied over time, such as weathering and rusting, are typically accepted as normal conditions.

This leaves us with two approaches.
The decision to choose weakly-supervised approach over semi-supervised approach was made due to the fact that precise annotation takes significantly longer time than approximate annotation. This is because most of the annotation time is spent on judging where the crack boundaries are, as the boundaries between crack and non-crack regions are often blurry and ambiguous. In addition, the boundaries are tedious to annotate, because boundary lines are often complex. With approximate annotation, the annotators can ignore both of these difficulties.
Therefore, we decided to formulate crack detection as a weakly-supervised problem.

\subsection{Two Branched Approach} \label{subsec_framework}

Formulating crack detection as a weakly-supervised problem means that the annotation labels contain mistakes, mostly around the crack boundaries. Therefore, a crack detector must be able to recover the boundary information lost during the annotation process. To do so, we observed how human annotators annotate cracks, and noticed that the annotation process is typically done in two steps. First, an annotator determines the rough locations of cracks by looking at the entire image. Then, the annotator zooms into a section, compares its pixel darkness against the neighbors, and annotates dark pixels as cracks.
To emulate these steps, we propose our two branched framework as shown in Fig. \ref{overall_framework}.

First, the \textit{Macro Branch} is a supervised branch trained using the rough annotations, and it determines the rough locations of the cracks.
On the other hand, the \textit{Micro Branch}'s task is to recover the precise boundary information from localized information. In our implementation, this branch conducts a simple per-pixel darkness calculation.
The outputs of two branches are fused by a multiplication, which is then thresholded to produce the final output.
Because any traditional crack detector can take the role of the Macro Branch without any modification, this framework can easily be integrated to any existing crack detection systems by simply adding the Micro Branch. This also implies that our framework can take advantages of future developments in fully-supervised crack detectors.

At first glance, the proposed framework may appear similar to the feature fusion technique, in which features extracted from various models are combined to make the final prediction. 
In fact, the proposed method can be viewed as a special case of feature fusion, where the two branches extract information from local and global contexts. 
However, there is a key difference in that the two branches are trained separately and their outputs are combined after probability predictions, unlike in typical feature fusion where the features are fused before the final prediction and the supervisory signal is back-propagated through the whole system. This is because in weakly-supervised setting, the final goal is not to emulate the low quality annotation, but to predict a more refined version of it. So directly training the Micro Branch with the weakly-supervised annotation is often harmful. Therefore, the fusion occurs after probabilities are calculated as opposed to after features are calculated.

Note that the proposed Micro Branch implementation is very simple, and could easily be improved by adding more pre- and post-processes.
Minimal working configuration is consciously chosen to show the effectiveness of the framework design. Further explorations on the Micro Branch configurations are subjects of future research.

\begin{table}
	\centering
	\caption{Dataset information and annotation times (in hours). $^\ast$Indicates that only train images are annotated.}
	\begin{tabular}{@{\extracolsep{4pt}}cccccc@{}}
		\hline
		\multirow{2}{*}{Dataset}& \multicolumn{2}{c}{Sample counts} & \multicolumn{3}{c}{Annotation times (hr.)} \\
		\cline{2-3} \cline{4-6}
		                        &Train&Test&Precise   & Rough & Rougher \\\hline
		Aigle   & 24    & 14    & -     & 0.45          & 0.3 \\
		CFD     & 71    & 47    & 21.5  & 2.3           & 0.72 \\
		DCD     & 300   & 237   & -     & 8.1$^\ast$    & 1.45$^\ast$  \\ \hline
	\end{tabular}
	\label{dataset_overview}
\end{table}

\section{Experiments}

\subsection{Experimental Setup}

\subsubsection{Datasets}

Aigle dataset (Aigle) \cite{aiglern}, Crack Forest Dataset (CFD) \cite{cfd}, and DeepCrack Dataset (DCD) \cite{liu2019deepcrack} are used for evaluation (details summarized in Table \ref{dataset_overview}). In order to distinguish the original annotations from the others, the original annotations will be referred to as the Precise Annotation.

\subsubsection{Evaluation Metrics}

Complying with the previous crack detection literature, F1-score, which is the harmonic mean of precision and recall, is chosen as the evaluation metric. More specifically, the Macro F1-score\footnote{In this paper, the Macro F1-score is defined as the harmonic mean of average precision and average recall values.} is evaluated because it weighs all images equally toward the overall score, unlike the Micro F1-score which weighs images with thicker cracks more. Also, pixel tolerances, typically introduced to absorb the annotation inaccuracies \cite{fan_automatic_2018, inoue}, are not adopted because it ambiguates the effect of low quality annotations.

\begin{table}
	\centering
	\setlength{\tabcolsep}{4pt}
    \renewcommand{\arraystretch}{1.1}
	\caption{Training parameters of the tested models. Inoue Light model shares the same parameters as Inoue \etal.}
	\begin{tabular}{lcccl}
		\hline
		Model       & lr    & \# epochs & $w_c^{model}$ & Notes \\ \hline
		Inoue \etal & 1e-1  & 50        & 20  	        & lr halved every 20 epochs \\ \hline
		DeepCrack   & 1e-3  & 700       & 33.3          & lr decayed after 400 epochs \\ \hline
		\multirow{2}{*}{DeepLab v3+} & \multirow{2}{*}{5e-3}  & \multirow{2}{*}{200000}    & \multirow{2}{*}{100}           & Xception 65 backbone \cite{chollet2017xception}, \\
		            &       &           &               & PASCAL VOC pretrained \\ \hline
	\end{tabular}
	\label{model_settings}
\end{table}

\subsection{Macro Branch Models} \label{model_desc}

To assess the versatility of our proposed framework, it was tested with four semantic segmentation model architectures.
First, DeepCrack by Liu \etal \cite{liu2019deepcrack}\footnote{Implementation adopted from https://github.com/yhlleo/DeepSegmentor} and the model proposed by Inoue \etal \cite{inoue}, which are both architectures designed specifically for crack detection, are selected to assess the effectiveness of the proposed framework for task-specific models. In addition, DeepLab V3+ \cite{deeplabv3p}\footnote{Implementation adopted from https://github.com/tensorflow/models/}, which is designed for general semantic segmentation tasks, is selected to evaluate the proposed framework on general task models. Finally, a computationally lighter version of Inoue \etal's model is also selected (named Inoue Light) to evaluate the framework on extremely light models. In Inoue Light models, \textit{CONV 2} and \textit{CONV 4} layers have strides of 2 instead of 1, and the number of filters at each layer are halved.
Experimental parameters of the above models are summarized in Table \ref{model_settings}.

\subsection{Low Quality Annotation Preparation} \label{lq_anno}

In order to evaluate the proposed framework under weakly-supervised setting, two annotations of different qualities are manually annotated (Rough and Rougher Annotations).
The time taken for annotation are summarized in Table \ref{dataset_overview}. Note that the annotation times for the Precise Annotation are unknown, so we reannotated CFD precisely to approximate the value. As the table shows, the annotation times required for the Rough and Rougher Annotations are an order of magnitude shorter than that of the Precise Annotation.
This verifies our choice of weakly-supervised learning over semi-supervised learning, as the time devoted to annotate the Rougher Annotations for CFD is only enough to precisely annotate 4 images, which is likely to be too small for effective training.

\begin{figure*}[!tb]
    \small
    \centering
    \renewcommand{\arraystretch}{0.6}
    \setlength{\tabcolsep}{1pt}
    \newcolumntype{C}{>{\centering\arraybackslash} m{2.6cm} }
    \newcolumntype{B}{>{\centering\arraybackslash} m{0.4cm} }
	\begin{tabular}{BCCCCCC}

    \rotatebox[origin=c]{90}{Aigle} &
    \includegraphics[width=\linewidth]{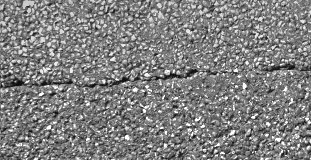} &
    \includegraphics[width=\linewidth]{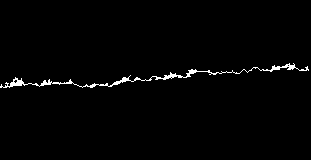} &
    \includegraphics[width=\linewidth]{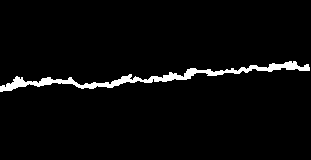} &
    \includegraphics[width=\linewidth]{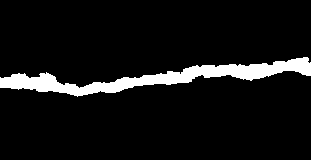} &
    \includegraphics[width=\linewidth]{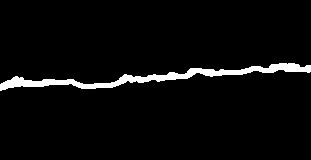} &
    \includegraphics[width=\linewidth]{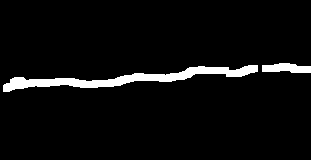} \\

    \rotatebox[origin=c]{90}{CFD} &
    \includegraphics[width=\linewidth]{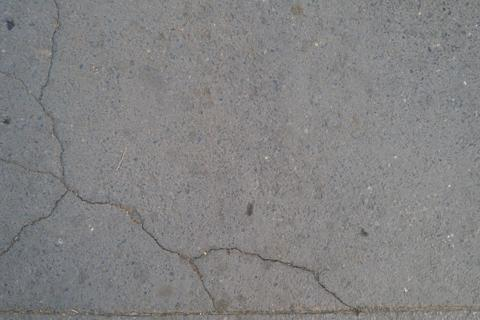} &
    \includegraphics[width=\linewidth]{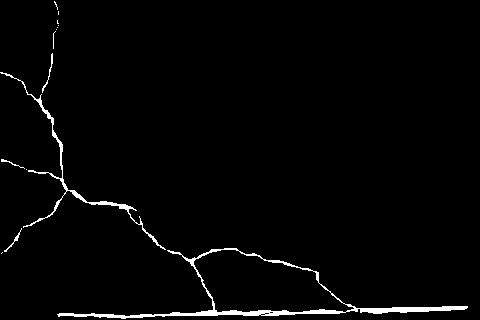} &
    \includegraphics[width=\linewidth]{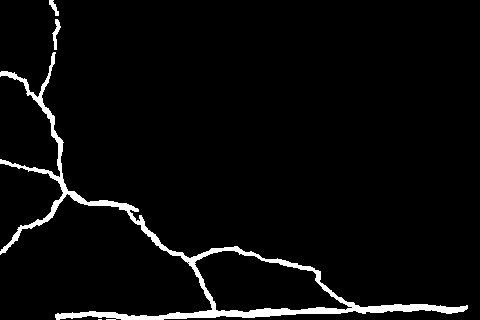} &
    \includegraphics[width=\linewidth]{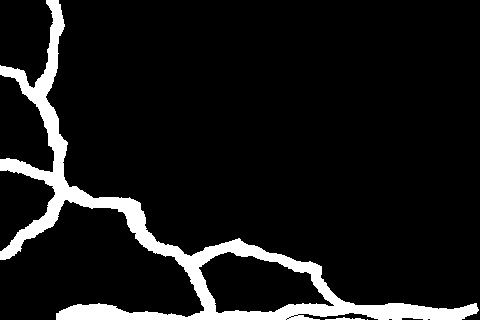} &
    \includegraphics[width=\linewidth]{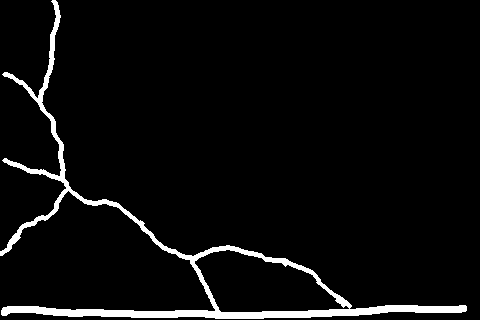} &
    \includegraphics[width=\linewidth]{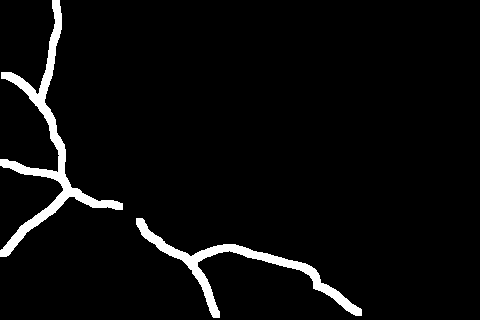} \\

    \rotatebox[origin=c]{90}{DCD} &
    \includegraphics[width=\linewidth]{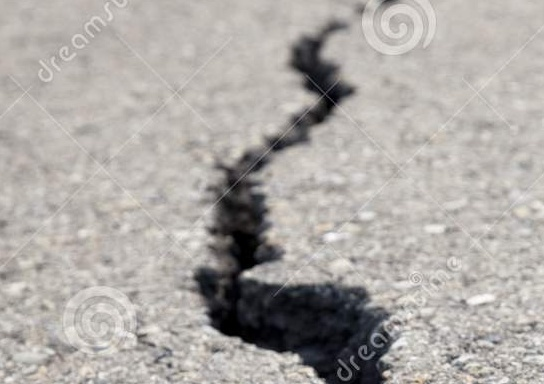} &
    \includegraphics[width=\linewidth]{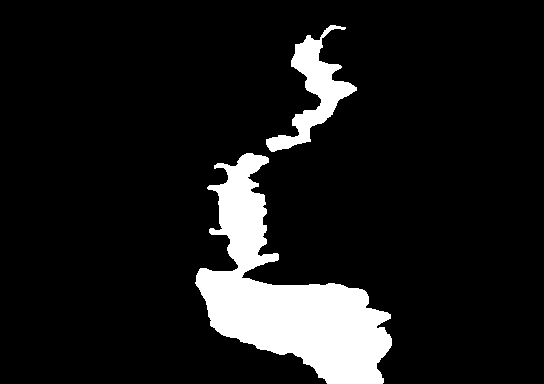} &
    \includegraphics[width=\linewidth]{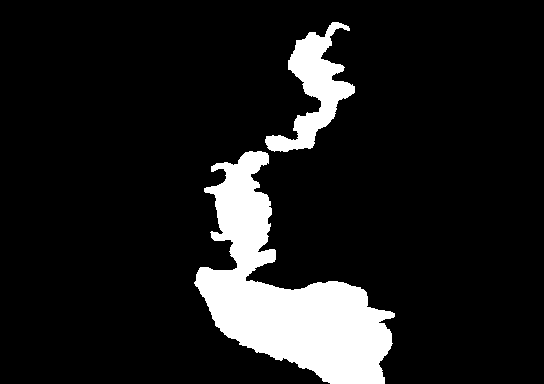} &
    \includegraphics[width=\linewidth]{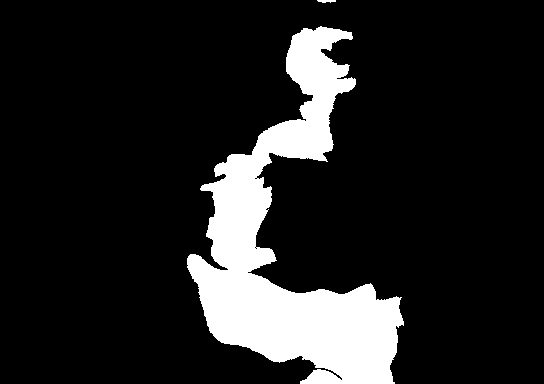} &
    \includegraphics[width=\linewidth]{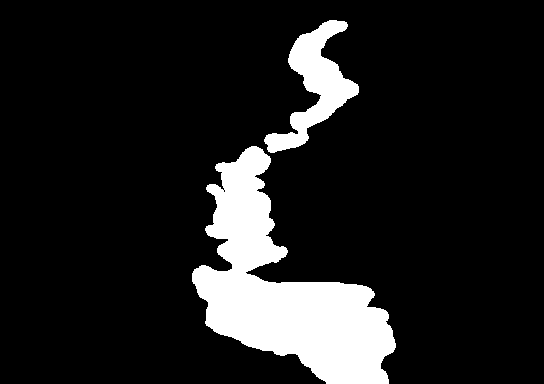} &
    \includegraphics[width=\linewidth]{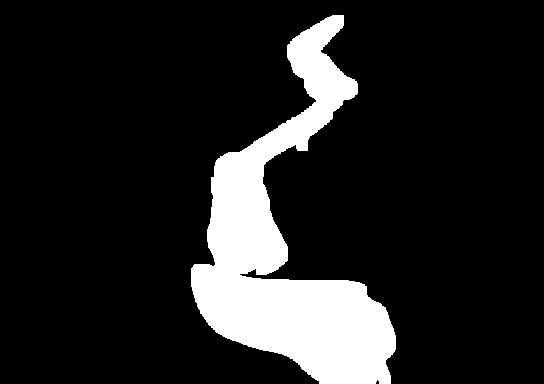} \\

    \\[-1.5mm]
     & Original & Precise & Dilation 1 & Dilation 4 & Rough & Rougher \\

    \end{tabular}
    \caption{Annotation samples (white pixels: crack regions, black pixels: non-crack regions).}
    \label{gt_samples}
\end{figure*}

\begin{figure}[!tb]
    \small
    \centering
    \renewcommand{\arraystretch}{0.6}
    \setlength{\tabcolsep}{1pt}
    \newcolumntype{C}{>{\centering\arraybackslash} m{2.9cm} }
	\begin{tabular}{CCC}

    \includegraphics[width=\linewidth]{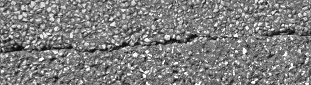} &
    \includegraphics[width=\linewidth]{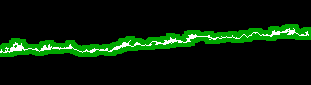} &
    \includegraphics[width=\linewidth]{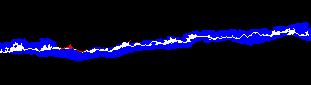} \\

    \includegraphics[width=\linewidth]{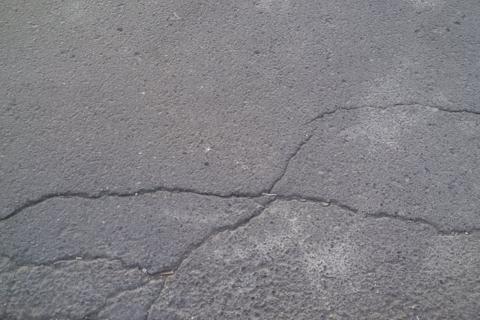} &
    \includegraphics[width=\linewidth]{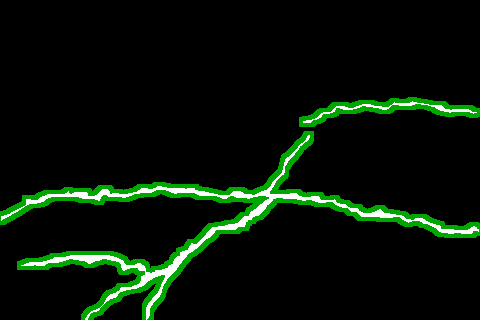} &
    \includegraphics[width=\linewidth]{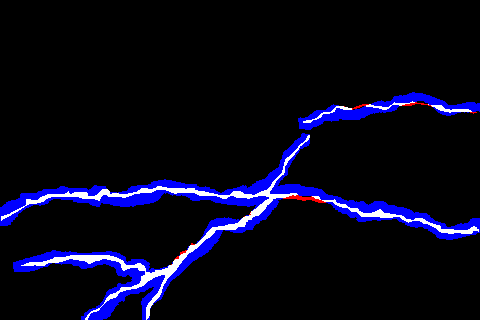} \\

    \includegraphics[width=\linewidth]{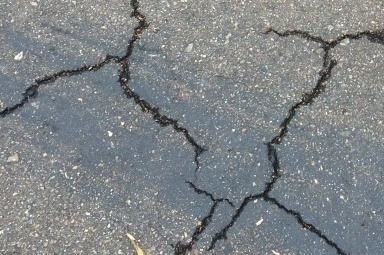} &
    \includegraphics[width=\linewidth]{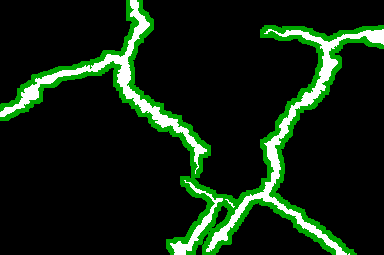} &
    \includegraphics[width=\linewidth]{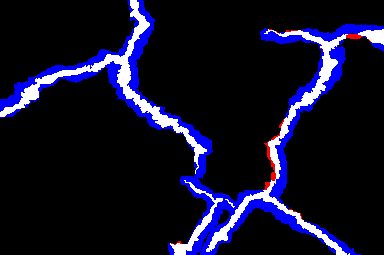} \\

    \\[-1.5mm]
    Input & w/o Elastic Transform & w/ Elastic Transform \\

    \end{tabular}
    \caption{Dilation Annotation synthesis process for $n_{dil}$=4. Best viewed in color. White: Precise Annotation, Green: dilation only, Blue: with Elastic Transform (\ie the Dilation Annotation), Red: regions missed by the Dilation Annotation.}
    \label{dil_samples}
\end{figure}

In addition to manual annotations, a suite of synthetic annotations are prepared to better control and quantify the annotation quality, and also to obtain annotations of arbitrary quality at a small cost.
To create synthetic annotations, we start by applying the dilation operation to the Precise Annotation $n_{dil}$ times in order to emulate the over-annotating nature of the low quality annotation. However, using this dilation result as the synthetic low quality annotation has a major problem, as the true crack region is always located at the center of the dilated crack region, and also the edges of the dilated crack region perfectly matches that of the true crack region. To fix this problem, we apply Elastic Transform \cite{simard2003best} to the dilation result, to "morph" parts of the dilated results. In addition, when applying Elastic Transform, we ensure that the synthesized annotation's crack regions cover between 92.5\% to 97.5\% of the true crack pixels, so that the annotations are not severely morphed. Setting the target coverage value to be less than 100\% is intentionally done to simulate rushed human annotators who are likely to miss parts of cracks during annotation.

To test the proposed framework under various settings, four annotations with different $n_{dil}$ are synthesized ($n_{dil} \in \{1,2,3,4\}$).
The four synthesized annotations will be referred to using $n_{dil}$ values used during the synthesis process. In other words, the annotation quality decreases from Dilation 1 Annotation to Dilation 4 Annotation. Please refer to Fig. \ref{gt_samples} for samples of both manual and synthesized annotations, and Fig. \ref{dil_samples} for sample synthesis processes for $n_{dil}=4$.
For more details on the annotation preparation process, please refer to Sec. \ref{supp:manual} and \ref{supp:synth} of the Supplementary Materials.

\subsection{Evaluation on the Low Quality Annotations} \label{sec_eval_lq}

\begin{figure}[!tb]
	\centering
	\includegraphics[width=\linewidth]{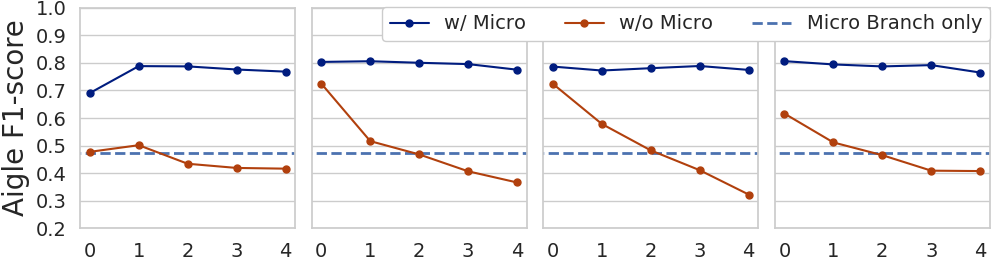} \\
	\vspace{1mm}
	\includegraphics[width=\linewidth]{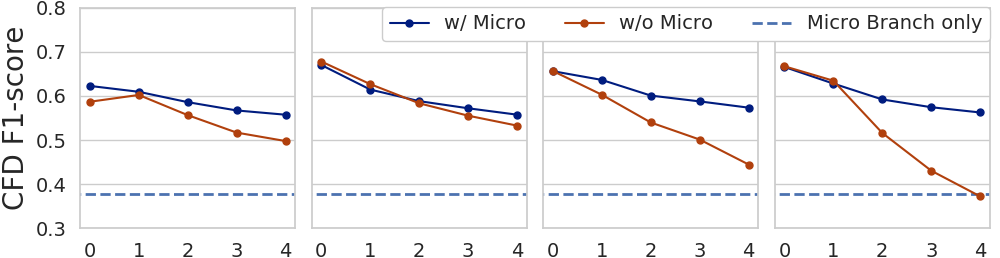} \\
	\vspace{1mm}
	\includegraphics[width=\linewidth]{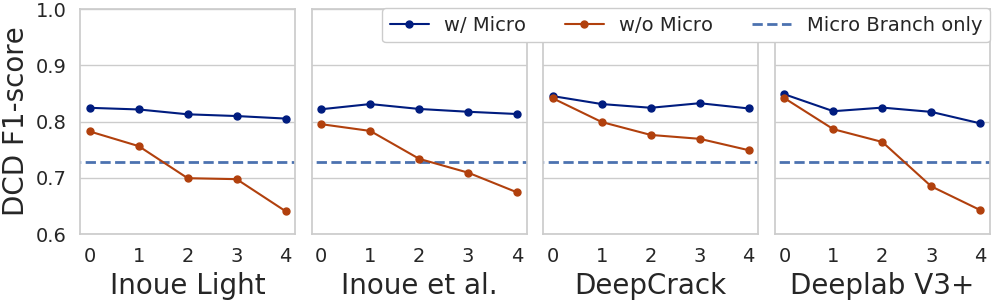}
	\caption{Results on the Dilation Annotation. Horizontal axis: $n_{dil}$ used to synthesize the annotation.}
	\label{plot_dil}
\end{figure}

\begin{figure}[!tb]
	\centering
	\includegraphics[width=\linewidth]{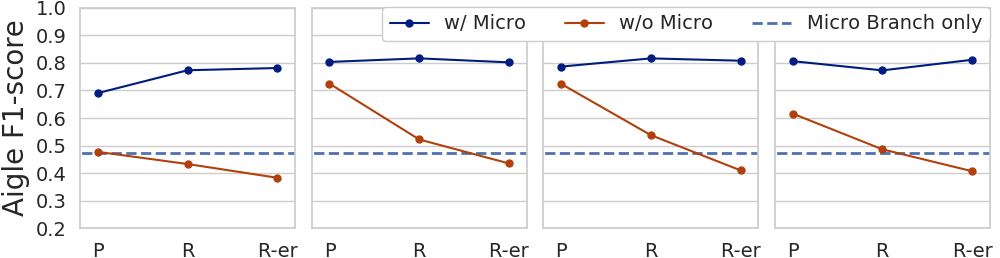} \\
	\vspace{1mm}
	\includegraphics[width=\linewidth]{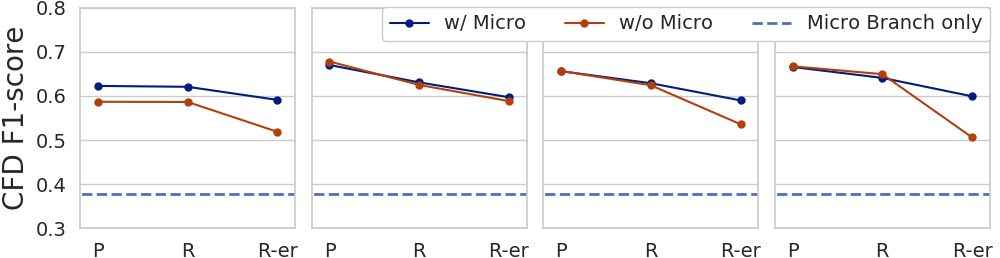} \\
	\vspace{1mm}
	\includegraphics[width=\linewidth]{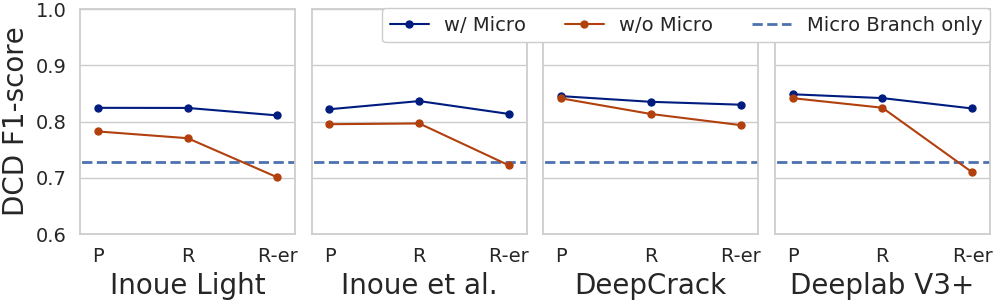}
	\caption{Results on the Precise (P),  Rough (R), and Rougher (R-er) Annotations.}
	\label{plot_rough}
\end{figure}

Fig. \ref{plot_dil} and \ref{plot_rough} show the results of the models trained with the low quality annotations. As the graphs show, the proposed framework (blue solid lines) better retains the performance compared to the baseline Macro Branch only case (orange lines) for most settings. This demonstrates the effectiveness of the proposed framework, for both manual and synthetic annotations. In fact, for Aigle, the result improves for annotations of lower quality.
On the other hand, the proposed framework performs poorly on CFD, dropping from the Precise Annotation result by 11.2\% for the synthetic annotations and 7.3\% for the manual annotations in the worst case.

\begin{figure}[!tb]
	\centering
	\includegraphics[width=1\linewidth]{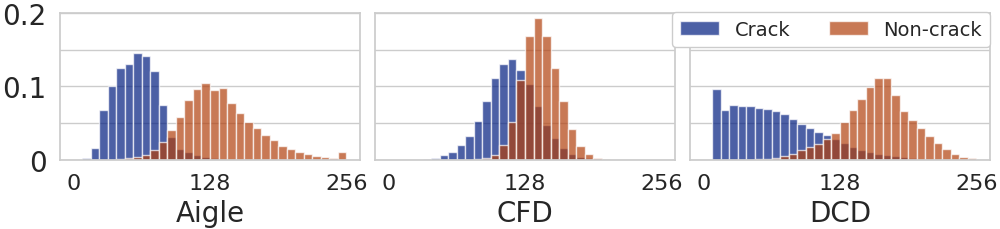}
	\caption{Brightness distribution for crack and non-crack regions. Horizontal axis: pixel brightness value.}
	\label{brightness_hist}
\end{figure}

The reason for this variation in performance across different datasets lies in the validity of the assumption that crack pixels are dark (\textit{brightness assumption}). This assumption is critical for the success of the darkness calculation strategy taken by the Micro Branch.
By plotting the brightness distributions of the crack and non-crack regions as shown in Fig. \ref{brightness_hist}, it becomes clear that for CFD, brightness is not a strong characteristic that distinguishes the two regions. This suggests that the brightness assumption is faulty for CFD, explaining the low performance.

On the other hand, the brightness distributions are distinctly separated for Aigle and DCD. In fact, the assumption holds so strongly for DCD that an F1-score of 72.8\% can be achieved on DCD just by using the pixel darkness as the crack probability. When the brightness assumption holds true, it means that the Micro Branch can better refine the Macro Branch output. Therefore, it becomes more important for the Macro Branch to obtain higher recall through over-detection. As lower quality annotations encourage the Macro Branch to over-detect, this explains why the framework can successfully maintain high accuracy for lower quality annotations of Aigle and DCD, sometimes even surpassing the results of the Precise Annotation.

\begin{table}
	\centering
	\caption{Maximum F1-score difference between the four models tested, for various annotation strategies and datasets. "MiB" column indicates the inclusion of the Micro Branch in the inference process.}

    \newcolumntype{D}{>{\arraybackslash} m{0mm}}
    \newcolumntype{E}{>{\centering\arraybackslash} m{3mm}}
    \newcolumntype{C}{>{\centering\arraybackslash} m{1cm}}
    \renewcommand{\arraystretch}{1.3}
	\begin{tabular}{Dc|ccccc}
		\hline
		&\multirow{2}{*}{MiB}   & \multicolumn{5}{c}{Maximum Difference} \\
		&                       & Precise   & Rough & Rougher   &Dilation 1  &Dilation 4  \\\hline
		\multirow{2}{*}{\rotatebox[origin=c]{90}{Aigle}}    &   \checkmark  &0.116&0.044&0.030&0.034&0.010 \\
		                                                    &               &0.247&0.105&0.052&0.077&0.094 \\\hline
		\multirow{2}{*}{\rotatebox[origin=c]{90}{CFD}}      &   \checkmark  &0.047&0.020&0.009&0.027&0.016 \\
		                                                    &               &0.091&0.063&0.082&0.032&0.160 \\\hline
		\multirow{2}{*}{\rotatebox[origin=c]{90}{DCD}}      &   \checkmark  &0.027&0.017&0.019&0.013&0.027 \\
		                                                    &               &0.059&0.054&0.093&0.043&0.109 \\\hline
	\end{tabular}

	\label{table_best_f1}
\end{table}

To show the strength of using the pixel darkness for crack detection, blue dotted lines are drawn in Fig. \ref{plot_dil} and \ref{plot_rough} to indicate the evaluation results of the case in which only the Micro Branch is used to detect cracks. As expected from Fig. \ref{brightness_hist}, the simple darkness calculation performs particularly well for Aigle and DCD, even outperforming the Macro Branch only models in some cases.
Also, the figures show that the results of the proposed framework is always better than applying the Micro Branch alone, implying that the Macro and the Micro Branches complement each other.

By comparing the model performances in Fig. \ref{plot_dil} and \ref{plot_rough}, we can approximate the annotation quality of the Rough and Rougher Annotations to equal that of the Dilation 1 and Dilation 2 Annotations, respectively. 
This suggests that if one can tolerate the performance degradation observed for the Dilation 3 and 4 Annotations, they may want to opt for even lower quality (and therefore faster) annotation policies.

In Table \ref{table_best_f1}, we summarize the maximum F1-score difference among the four models tested, for various annotation strategies and datasets. Interestingly, in all cases tested, the inclusion of the Micro Branch reduces the maximum F1-score difference among the four models, implying that simpler models with poorer detection accuracy benefits more from the proposed framework.
This observation can be explained by analyzing the effect of the proposed framework on the complexity of the crack detection problem. Under the proposed framework, the supervised models only have to identify the rough locations of the cracks, as the precise detection details are taken care by the rule-based Micro Branch. We speculate that the proposed framework simplified the crack detection problem enough that having higher model complexity now yields diminishing return, explaining the reduced performance gap across the detection models as observed in Table \ref{table_best_f1}.

\begin{table*}
	\centering
	\caption{Model performance (F1-score) across datasets. Rows and columns are organized by test and train datasets, respectively. "MiB only" indicates the result of only applying the Micro Branch. \textbf{Bolded values} indicate the best results for the model, \underline{underlined values} indicate lower score than that of MiB only case.}
    \newcolumntype{B}{>{\centering\arraybackslash} m{0cm} }
    \newcolumntype{C}{>{\centering\arraybackslash} m{0.9cm} }
    \newcolumntype{D}{>{\centering\arraybackslash} m{1.2cm} }
    \newcolumntype{E}{>{\centering\arraybackslash} m{-5mm} }
    \setlength{\tabcolsep}{2pt}
	\begin{tabular}{@{\extracolsep{4pt}}BBccccEcccEcccEcccED@{}}
		\hline
		 & & & \multicolumn{15}{c}{Train Dataset} \\
		 & & & \multicolumn{3}{c}{Inoue Light} && \multicolumn{3}{c}{Inoue} && \multicolumn{3}{c}{DeepCrack} && \multicolumn{3}{c}{DeepLab V3+} &&\multirow{2}{*}{MiB only} \\
		\cline{4-6} \cline{8-10} \cline{12-14} \cline{16-18}
    	    &&       &Aigle  &CFD    &DCD    &&Aigle  &CFD    &DCD    &&Aigle  &CFD    &DCD    &&Aigle  &CFD    &DCD    &    \\
		\multirow{3}{*}{\rotatebox[origin=c]{90}{Test}} &\multirow{3}{*}{\rotatebox[origin=c]{90}{Dataset}}
		     &Aigle &\textbf{0.477}  &\underline{0.461}  &\underline{0.405}  &&\textbf{0.725}  &0.509  &0.501  &&\textbf{0.724}  &0.534  &0.571  &&\textbf{0.616}  &\underline{0.281}  &\underline{0.444}    &&0.473  \\
		    &&CFD   &0.571  &\textbf{0.587}  &0.525  &&0.638  &\textbf{0.678}  &0.603  &&0.637  &\textbf{0.657}  &0.566  &&0.583  &\textbf{0.667}  &0.556    &&0.377  \\
		    &&DCD   &\underline{0.659}  &\underline{0.674}  &\textbf{0.782}  &&0.764  &\underline{0.727}  &\textbf{0.795}  &&\underline{0.548}  &\underline{0.330}  &\textbf{0.841}  &&0.819  &\underline{0.669}  &\textbf{0.842}    &&0.728  \\ \hline
	\end{tabular}
	\label{table_across_dsets}
\end{table*}

\begin{table*}
	\centering
	\caption{Comparisons between out-of-domain results and weakly-supervised results. R and R-er refer to Rough and Rougher Annotations, respectively. Table organization same as Table \ref{table_across_dsets}. \textbf{Bolded values} indicate the best results for a model.}

    \newcolumntype{B}{>{\centering\arraybackslash} m{0cm} }
    \newcolumntype{D}{>{\centering\arraybackslash} m{0.45cm}}
    \newcolumntype{C}{>{\centering\arraybackslash} m{0.55cm}}
    \setlength{\tabcolsep}{2pt}
	\begin{tabular}{@{\extracolsep{4pt}}BDCCCCCCCCCCCCCCCCCCCC@{}}
		\hline
		 & & \multicolumn{20}{c}{Train} \\
		 & & \multicolumn{5}{c}{Inoue Light} & \multicolumn{5}{c}{Inoue} & \multicolumn{5}{c}{DeepCrack} & \multicolumn{5}{c}{DeepLab V3+} \\
		\cline{3-7} \cline{8-12} \cline{13-17} \cline{18-22}
    	    &      &Aigle  &CFD    &DCD    &R      &R-er   &Aigle  &CFD    &DCD    &R      &R-er   &Aigle  &CFD    &DCD    &R      &R-er   &Aigle  &CFD    &DCD   &R      &R-er   \\
		\multirow{3}{*}{\rotatebox[origin=c]{90}{Test}}
	     &Aigle &- &   0.645 &  0.475 &  0.773 & \textbf{0.781} &  - &   0.669 &  0.654 & \textbf{0.816} & 0.802 & - &   0.748 &  0.694 & \textbf{0.816} &   0.808 & - & 0.278 &  0.498 &  0.773 &    \textbf{0.811} \\
	     &CFD &0.607 &  - &  0.542 &  \textbf{0.621} & 0.591 &  \textbf{0.647} &  - &  0.625 & 0.631 & 0.597 &  \textbf{0.635} &  - & 0.548 &    0.629 &   0.590 & 0.590 &  - &  0.573 &  \textbf{0.641} &    0.599 \\
	     &DCD &0.721 &   0.760 &    - &  \textbf{0.824} & 0.811 &  0.795 &   0.785 &    - & \textbf{0.836} & 0.813 &  0.565 &   0.382 & - & \textbf{0.835} &   0.830 & 0.829 & 0.686 &  - &  \textbf{0.842} &    0.823 \\ \hline
	\end{tabular}
	\label{table_across_dsets_comp}
\end{table*}

\begin{figure*}[!tb]
    \small
    \centering
    \renewcommand{\arraystretch}{0.4}
    \setlength{\tabcolsep}{0.5pt}
    \newcolumntype{C}{>{\centering\arraybackslash} m{1.6cm}}
    \newcolumntype{B}{>{\centering\arraybackslash} m{0.7cm}}
    \newcolumntype{D}{>{\centering\arraybackslash} m{0.5mm}}
	\begin{tabular}{BCCDCCDCCDCCDCC}

    \rotatebox[origin=c]{90}{Precise} &
    \includegraphics[width=\linewidth]{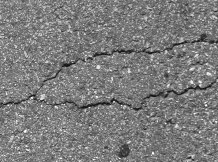} &
    \includegraphics[width=\linewidth]{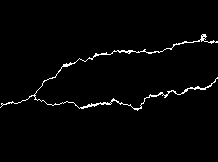} &&
    \includegraphics[width=\linewidth]{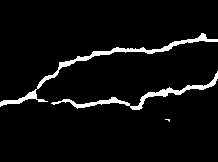} &
    \includegraphics[width=\linewidth]{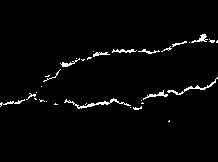} &&
    \includegraphics[width=\linewidth]{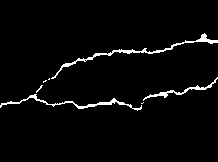} &
    \includegraphics[width=\linewidth]{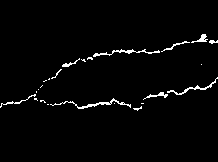} &&
    \includegraphics[width=\linewidth]{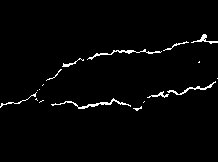} &
    \includegraphics[width=\linewidth]{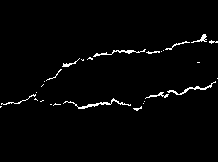} &&
    \includegraphics[width=\linewidth]{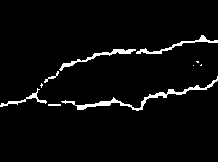} &
    \includegraphics[width=\linewidth]{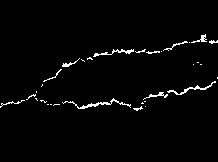} \\

    \rotatebox[origin=c]{90}{Dilation 1} &
    \includegraphics[width=\linewidth]{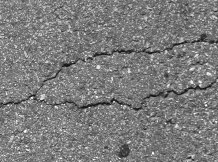} &
    \includegraphics[width=\linewidth]{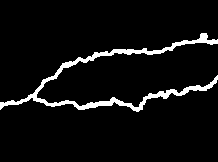} &&
    \includegraphics[width=\linewidth]{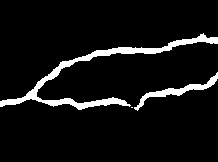} &
    \includegraphics[width=\linewidth]{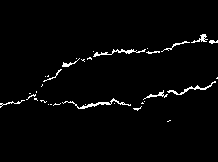} &&
    \includegraphics[width=\linewidth]{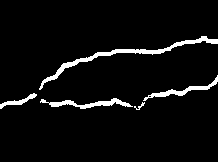} &
    \includegraphics[width=\linewidth]{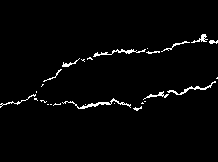} &&
    \includegraphics[width=\linewidth]{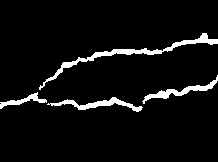} &
    \includegraphics[width=\linewidth]{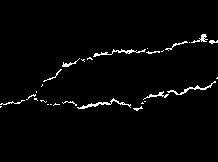} &&
    \includegraphics[width=\linewidth]{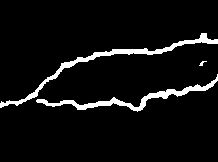} &
    \includegraphics[width=\linewidth]{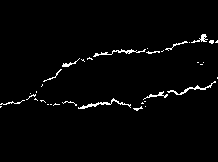} \\

    \rotatebox[origin=c]{90}{\parbox{1cm}{\centering Dilation 4}} &
    \includegraphics[width=\linewidth]{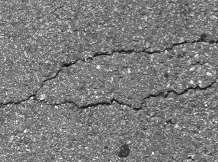} &
    \includegraphics[width=\linewidth]{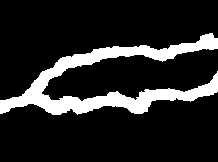} &&
    \includegraphics[width=\linewidth]{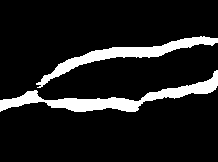} &
    \includegraphics[width=\linewidth]{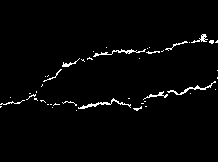} &&
    \includegraphics[width=\linewidth]{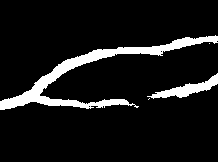} &
    \includegraphics[width=\linewidth]{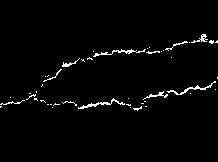} &&
    \includegraphics[width=\linewidth]{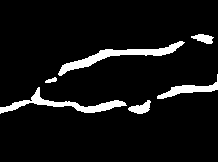} &
    \includegraphics[width=\linewidth]{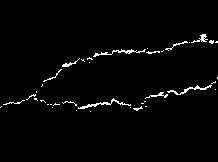} &&
    \includegraphics[width=\linewidth]{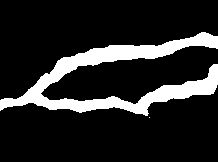} &
    \includegraphics[width=\linewidth]{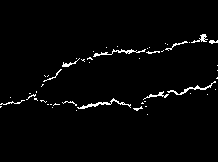} \\

    \rotatebox[origin=c]{90}{Rough} &
    \includegraphics[width=\linewidth]{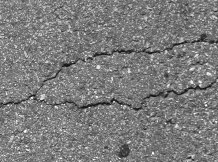} &
    \includegraphics[width=\linewidth]{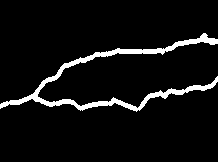} &&
    \includegraphics[width=\linewidth]{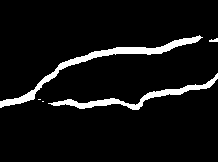} &
    \includegraphics[width=\linewidth]{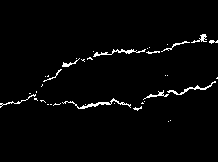} &&
    \includegraphics[width=\linewidth]{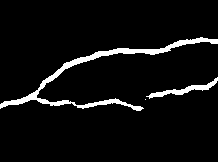} &
    \includegraphics[width=\linewidth]{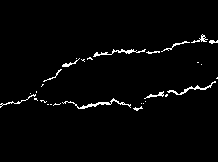} &&
    \includegraphics[width=\linewidth]{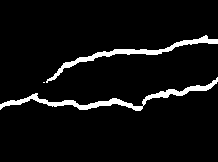} &
    \includegraphics[width=\linewidth]{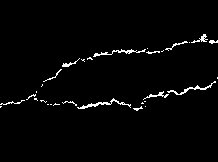} &&
    \includegraphics[width=\linewidth]{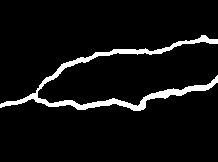} &
    \includegraphics[width=\linewidth]{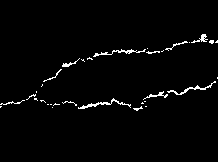} \\

    \rotatebox[origin=c]{90}{Rougher} &
    \includegraphics[width=\linewidth]{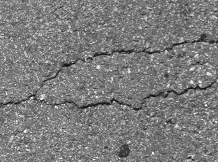} &
    \includegraphics[width=\linewidth]{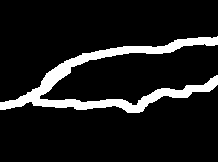} &&
    \includegraphics[width=\linewidth]{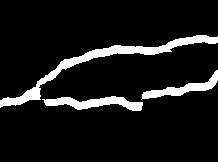} &
    \includegraphics[width=\linewidth]{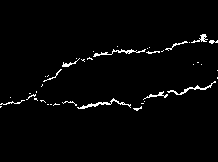} &&
    \includegraphics[width=\linewidth]{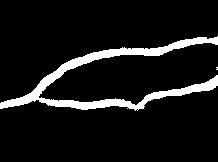} &
    \includegraphics[width=\linewidth]{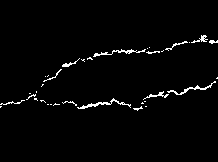} &&
    \includegraphics[width=\linewidth]{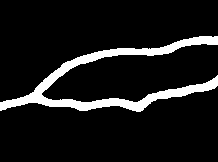} &
    \includegraphics[width=\linewidth]{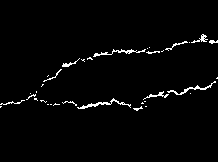} &&
    \includegraphics[width=\linewidth]{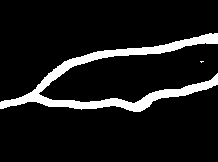} &
    \includegraphics[width=\linewidth]{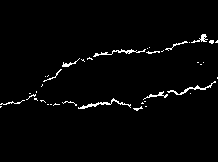} \\

    \\[-1.4mm]
     &\multirow{2}{*}{Original}&\multirow{2}{*}{Annotation}&   & w/o MiB& w/ MiB   &&w/o MiB& w/ MiB
     &&w/o MiB& w/ MiB  &&w/o MiB& w/ MiB \\

    \\[-1mm]
     &  &     && \multicolumn{2}{c}{Inoue \etal Light}    && \multicolumn{2}{c}{Inoue \etal}
     && \multicolumn{2}{c}{DeepCrack}         && \multicolumn{2}{c}{DeepLab V3+} \\

    \end{tabular}
    \caption{Inference examples (cropped for better viewing experience). MiB indicates including the Micro Branch in the inference.}
    \label{inf_samples}
\end{figure*}

Finally, the inference samples are shown in Fig. \ref{inf_samples}. As expected, models trained with less precise annotations learn to predict less precisely. Also, the inclusion of the Micro Branch greatly improves the prediction precision. However, one concern is that the Micro Branch outputs tend to be more fragmented, which may not be desirable for some applications. This arises from the Micro Branch predicting from highly localized information, so applying image smoothing to the Micro Branch input may relieve the problem.

\subsection{Comparing Weakly-Supervised Approach and Model Reuse}

One of the biggest motivations of this paper is the speculation that crack detectors do not generalize well across different domains. To evaluate this point, we trained the models on the Precise Annotation and tested them on different datasets. Results are summarized in Table \ref{table_across_dsets}.
As expected, the best performance is achieved when the test and train datasets match (in-domain case, \textit{ID}), outperforming the cases in which the test and train datasets do not match (out-of-domain case, \textit{OOD}) by 13.4\% on average across all datasets.

Furthermore, there are no winning strategies for successful domain generalization.
First, no dataset generalizes well. Results indicate that Aigle is the most robust train dataset, outperforming the other two datasets in 7 of the 8 OOD settings. However, even for Aigle, the F1-score drops by 7.9\% on average compared to the ID case, which is quite large. Also taking into account that the worst case F1-score drop is 29.3\%, none of the datasets are appealing option for training a model across domains. In addition, Aigle's performance goes against the popular belief that larger datasets generalize better, as Aigle has the fewest number of samples among the used datasets. Thus simply annotating more samples does not yield a more versatile dataset that trains robust models.

Furthermore, there is no model architecture that generalizes well, either. Inoue Light is the most consistent model, with the F1-score only dropping by 6.6\% on average compared to the ID case. However, this is a contrived result, as the Inoue Light model performs poorly to begin with, leaving less room for performance drop. The model that generalizes second best is Inoue \etal model at 10.9\% average drop, which is quite large. Considering that in the worst case this model dropped F1-score by 22.4\%, none of the model architectures are appealing for multiple domain inference.

Also shown in the table at far right is the result of applying the Micro Branch only, without the Macro Branch. Since the Micro Branch does not need annotations, it acts as a simple unsupervised baseline. Surprisingly, this simple baseline outperforms about half of the OOD results (in 10/24), showcasing the difficulties of OOD inference.

To compare the performance of the OOD inference and weakly-supervised approach, Table \ref{table_across_dsets_comp} summarizes the OOD results and the results of the proposed framework on the Rough and Rougher Annotations. Note that for the Rough and Rougher Annotations (\ie weakly-supervised settings), the train and test domains are matched. Also, to make comparisons fair, all results (including OOD inferences) are refined with the Micro Branch.
As the table shows, the proposed framework outperforms the OOD inference in 22 of 24 cases. In addition, note that the underperformance in both cases are relatively small (1.6\% and 0.6\%).

After evaluating the results of ID inference, OOD inference, and proposed framework, we conclude that
(1) prediction accuracy is significantly degraded when test and train domains do not match
(2) this degradation cannot be relieved by selecting specific dataset to train from or model architecture to train with, so it is difficult to train a model to be consistently accurate across different domains
(3) proposed weakly-supervised approach significantly outperforms the OOD inference.
As the annotation cost of the proposed approach is kept minimal, we believe the large gain in accuracy compared to the OOD inference make our proposal an appealing option when deploying crack detectors in multiple domains.

\begin{figure}[!tb]
	\centering
	\includegraphics[width=\linewidth]{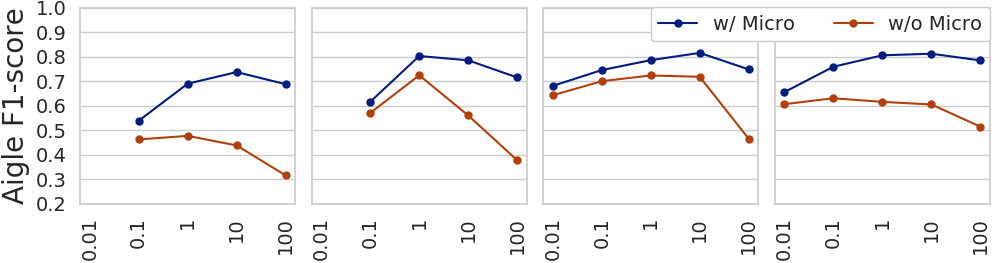} \\
	\vspace{1mm}
	\includegraphics[width=\linewidth]{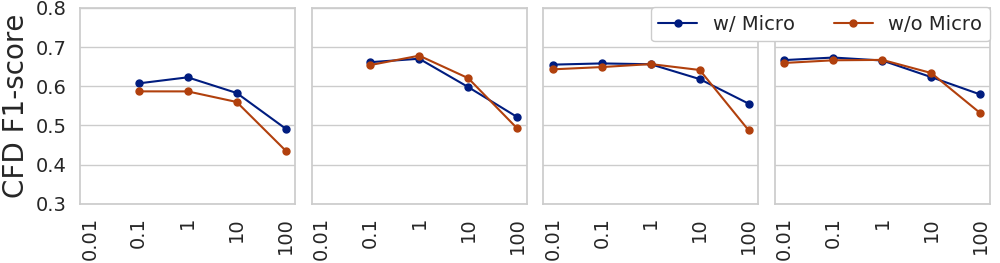} \\
	\vspace{1mm}
	\includegraphics[width=\linewidth]{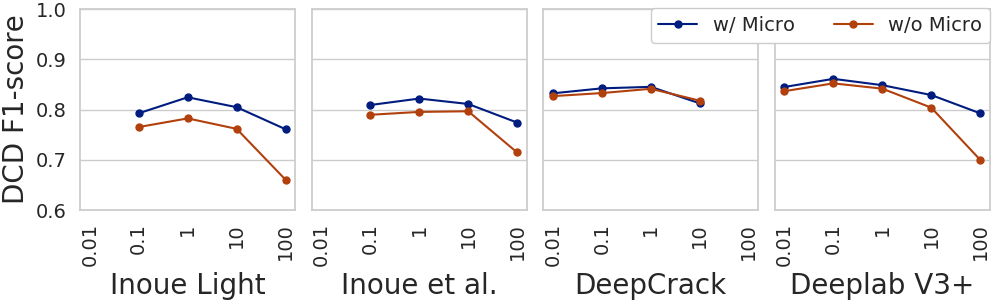}
	\caption{Effect of $w_c$ on the F1-score. Horizontal axis: $w_c$ used for training. Models are trained with the Precise Annotation.}
	\label{plot_weight}
\end{figure}

\begin{figure}[!tb]
	\centering
	\includegraphics[width=\linewidth]{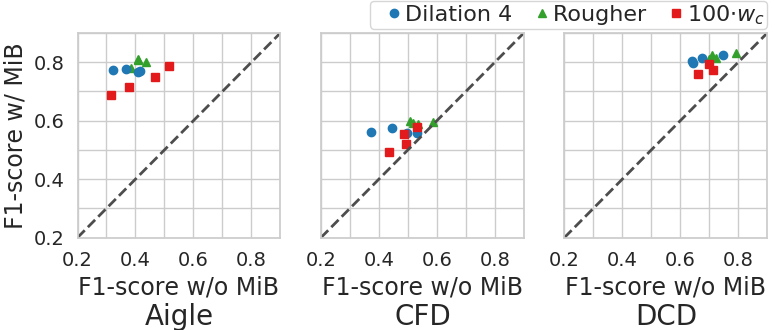}
	\caption{Effect of the Micro Branch refinement on F1-score. Each plotted point represents different training settings.}
	\label{fig_f1_improvement}
\end{figure}

\subsection{Class Rebalance Weight Sensitivity}

In previous sections we showed the effectiveness of the proposed framework in the weakly-supervised settings, and also compared against the case in which the same model is reused in multiple domains. From here, we explore another possibility and benefits of adopting the proposed framework.

A typical method for handling a class-imbalanced problem such as the crack detection problem is to add a class rebalance weight $w_c$ to the loss term so that the predictions on the crack pixels give more impact toward the weight update. Unfortunately, the optimal $w_c$ value ($w_c^{model}$) varies by models and sometimes by datasets.
However, one clear trend is that a higher $w_c$ encourages the model to over-detect. This is because higher $w_c$ intensifies the penalty for failing to detect the crack pixels, encouraging the model to predict any ambiguous regions as cracks.

As the inference samples in Fig. \ref{inf_samples} show, the Macro Branch trained with low quality annotations tend to over-detect, but the Micro Branch appropriately suppresses the over-detection. Under this observation, we hypothesized that the Micro Branch can also suppress over-detection caused by larger $w_c$, making models less sensitive against the values of $w_c$.

Fig. \ref{plot_weight} shows the result of training the models with the Precise Annotation under different values of $w_c$. The horizontal axis corresponds to the values of $w_c$, where 1 indicates $w_c=w_c^{model}$, 10 indicates $w_c=10 \cdot w_c^{model}$, and so forth. Please refer to Table \ref{model_settings} for specific values of $w_c=w_c^{model}$ used.
Note that some models failed to learn under extreme $w_c$, shown by blanks in the plot.
As hypothesized, the plots show that for Aigle and DCD, the inclusion of the Micro Branch significantly reduces the performance degradation caused by larger $w_c$. In fact, larger $w_c$ improved the detection accuracy for Aigle. This performance improvement due to over-detection is consistent with the previous observations in Sec. \ref{sec_eval_lq}.
On the other hand, improvements were small for smaller $w_c$, likely because the Micro Branch cannot correct under-detections.

However, the performance improvement was not as great as it was in Sec. \ref{sec_eval_lq}. To visualize this point, we plotted in Fig. \ref{fig_f1_improvement} the F1-scores for the Rougher, Dilation 4, and $100 \cdot w_c^{model}$ cases, with and without the Micro Branch. The inferior improvement for $100 \cdot w_c^{model}$ case is illustrated by the plot points corresponding to $100 \cdot w_c^{model}$ case (red squares) located lower in y-axis than the other two cases.

\begin{figure}[!tb]
    \small
    \centering
    \renewcommand{\arraystretch}{0.4}
    \setlength{\tabcolsep}{0.5pt}
    \newcolumntype{C}{>{\centering\arraybackslash} m{1.7cm} }
    \newcolumntype{B}{>{\centering\arraybackslash} m{0.3cm} }
	\begin{tabular}{CCCCC}
    
    \includegraphics[width=\linewidth]{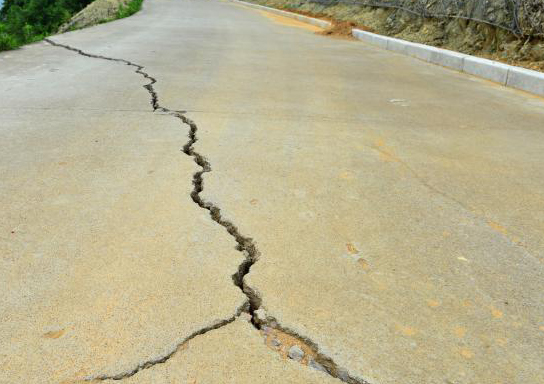} &
    \includegraphics[width=\linewidth]{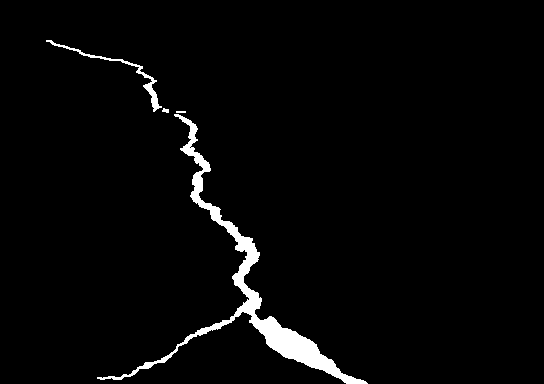} &
    \includegraphics[width=\linewidth]{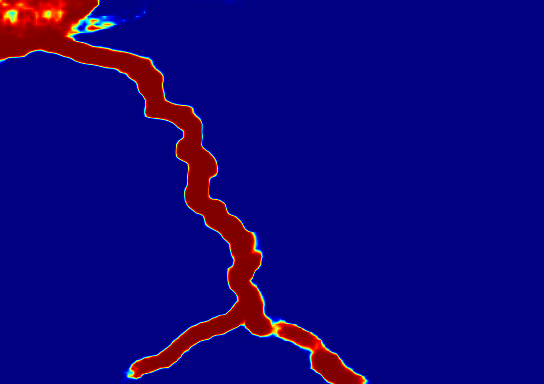} &
    \includegraphics[width=\linewidth]{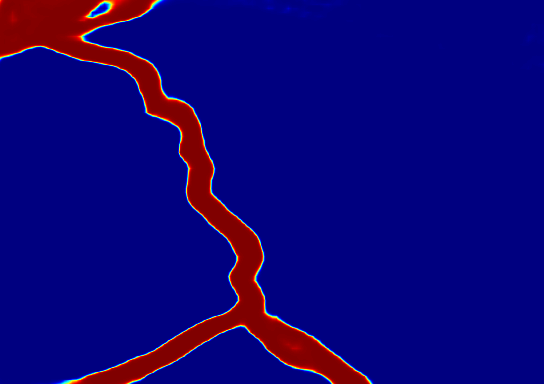} &
    \includegraphics[width=\linewidth]{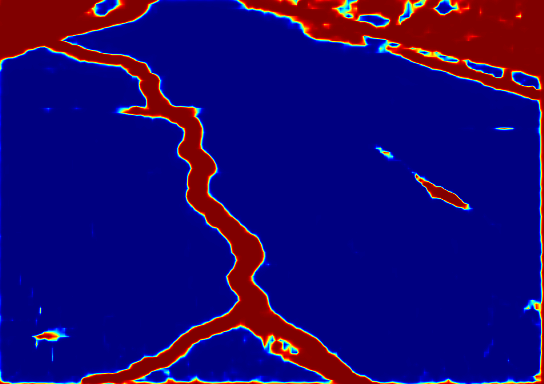} \\
    
    \includegraphics[width=\linewidth]{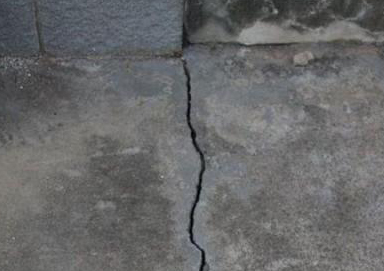} &
    \includegraphics[width=\linewidth]{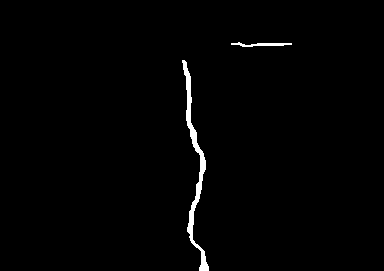} &
    \includegraphics[width=\linewidth]{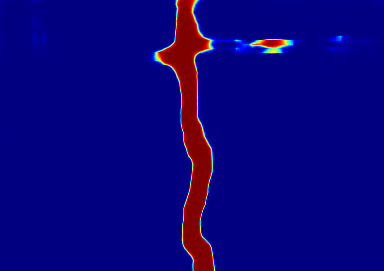} &
    \includegraphics[width=\linewidth]{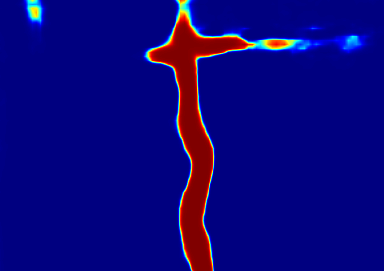} &
    \includegraphics[width=\linewidth]{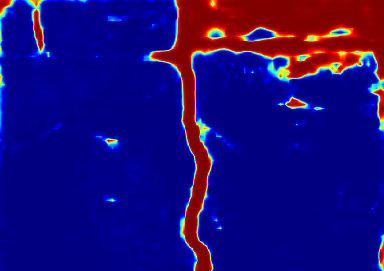} \\
    
    \includegraphics[width=\linewidth]{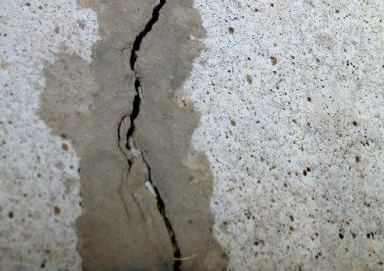} &
    \includegraphics[width=\linewidth]{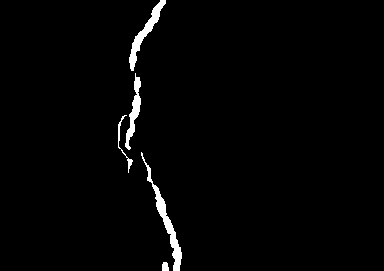} &
    \includegraphics[width=\linewidth]{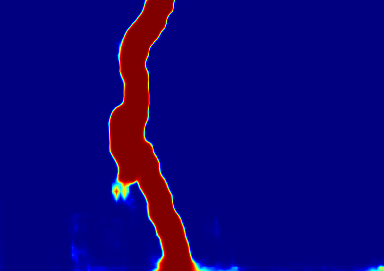} &
    \includegraphics[width=\linewidth]{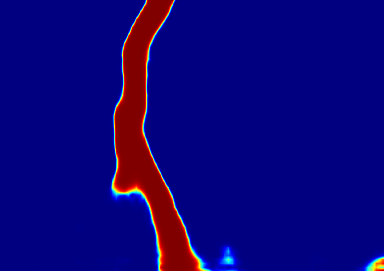} &
    \includegraphics[width=\linewidth]{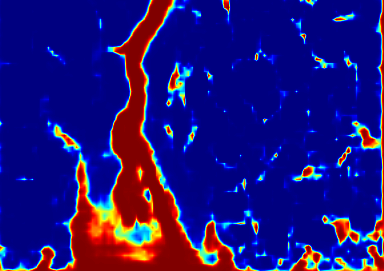} \\

    Original & GT & Dilation 4 & Rougher & 100$\cdot w_c^{model}$ \\

    \end{tabular}
    \caption{Sample heatmaps of Macro Branch predictions. Red indicates higher crack probability prediction.}
    \label{sample_hi_w_output}
\end{figure}

To investigate this inferior performance, we looked at the output samples as shown in Fig. \ref{sample_hi_w_output}, and noticed that $100 \cdot w_c^{model}$ models tend to over-detect dark regions even if they do not contain cracks at all. This is undesirable, as for the proposed framework to work optimally, the Macro Branch needs to locate the crack locations without heavily depending on the fact that cracks are dark. Since the Micro Branch makes decisions based solely on the pixel darkness values, if the Macro Branch's decision is also based on the pixel intensity, the Micro Branch cannot further refine the result.

Nevertheless, it is true that the proposed framework is able to correct for the over-detection caused by high $w_c$. Therefore, we conclude that the proposed framework has an added benefit that it is less sensitive to the class rebalance weight.

\section{Conclusion}

Automated inspection is increasingly becoming more important, as the number of roads and buildings in the world continually grows. However, current crack detectors require a large amount of cost-intensive annotations at each site in order to maintain high accuracy.
In this paper, we explored the weakly-supervised approach to the crack detection problem to reduce the annotation bottleneck, and proposed a framework that can retain high performance even when only low quality annotations are available. Our framework introduces a simple unsupervised component based on pixel brightness that can be added to any semantic segmentation networks to extend their ability to learn from low quality data.

Through thorough experiments on four different model architectures trained with manually and synthetically created low quality annotations of three different datasets, we showed that our framework is not only easily implementable but also highly effective under weakly-supervised settings, making it an attractive extension to existing crack detectors.

In the future, we would like to investigate alternative configurations for the proposed branches, and also make the Micro Branch more general, so it can be applied to other semantic segmentation tasks.



%

{\small
\bibliographystyle{unsrt}
\bibliography{references}
}

\newpage

\section{Supplementary Materials}

\subsection{Annotation Procedures for the Manual Annotations} \label{supp:manual}

In Sec. \ref{lq_anno}, we briefly discussed the manual annotations prepared to test the models under weakly-supervised settings. In this section, we describe the annotation rules that the annotators followed during the creation of the manual annotations. The annotation rules for the two manual annotations, named Rough and Rougher Annotations, are as follows (values in \{\} correspond to rules for the Rough Annotation, and [] correspond to rules for the Rougher Annotation):

Size of the pen used during annotation\footnote[1]{Note that all annotations are annotated with a pen tool in a paint application} was set to be \{1 or 2\} [3 or 4] pixels larger than the average width of the cracks, adjusted \{as many times as needed\} [once at most] per image. The cracks were traced in one stroke unless the crack width is larger than the pen size, and the small boundary contours were \{followed as much as possible\} [largely ignored].
For thick cracks of 8 pixels or wider, pen size of \{4\} [8] pixels was used to trace the outline. The outlined regions were filled afterwards.

\subsection{Annotation Procedures for the Synthetic Annotations} \label{supp:synth}

In Sec. \ref{lq_anno}, we briefly discussed the synthetic annotations prepared to test the models under weakly-supervised settings. In this section, we delve into the details of the synthesis pipeline, so that readers can better reproduce the dataset.

\begin{figure}[!tb]
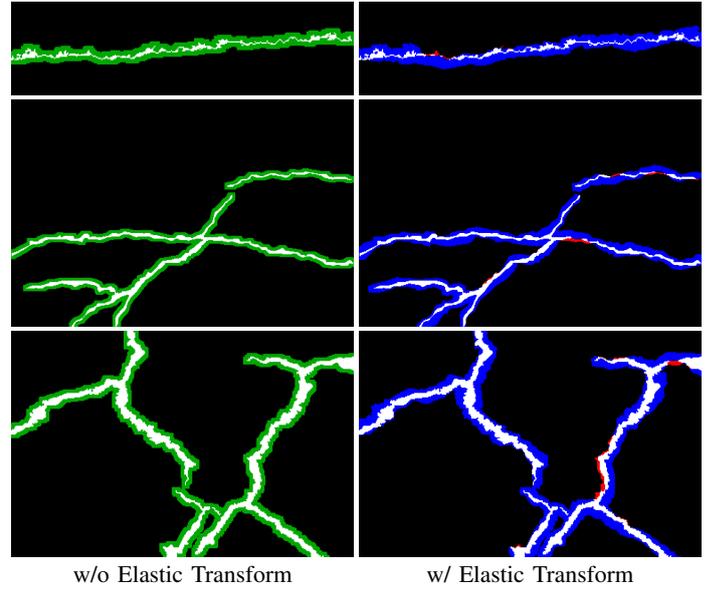

    \small
    \centering
    \renewcommand{\arraystretch}{0.6}
    \setlength{\tabcolsep}{1pt}
    \newcolumntype{C}{>{\centering\arraybackslash} m{4.5cm} }
	\begin{tabular}{CC}

    \includegraphics[width=\linewidth]{dilation_dataset/aigle_c18bor_2/dil4_combined.png} &
    \includegraphics[width=\linewidth]{dilation_dataset/aigle_c18bor_2/distort_combined2.png} \\

    \includegraphics[width=\linewidth]{dilation_dataset/cfd_02/dil4_combined.png} &
    \includegraphics[width=\linewidth]{dilation_dataset/cfd_02/distort_combined2.png} \\

    \includegraphics[width=\linewidth]{dilation_dataset/11125-1_2/dil4_combined.png} &
    \includegraphics[width=\linewidth]{dilation_dataset/11125-1_2/distort_combined2.png} \\

    \\[-1.5mm]
    w/o Elastic Transform & w/ Elastic Transform \\

    \end{tabular}
    \caption{Dilation Annotation synthesis process (for dilation=4). White: Precise Annotation, Green: dilation only, Blue: Elastic Transform applied to the dilation result (\ie new Dilation Annotation), Red: regions missed by the Dilation Annotation}
    \label{dil_samples2}
\end{figure}

The details of the synthesis pipeline is outlined in Algorithm \ref{synthesis_algo}.
We will describe the key steps here. 
First, a Precise Annotation sample $p$ is dilated $n_{dil}$ times to generate a dilated sample $d$ (steps 5-6). This $n_{dil}$ value dictates the quality of the resulting annotation. Examples of $p$ and $d$ are shown in the left column of Fig. \ref{dil_samples2} as white and green regions, respectively.
Then, Elastic Transform \cite{simard2003best} is applied to the dilated sample to form a synthesis candidate $s$ (step 9). An implementation by the popular data augmentation library Albumentations \cite{albumentations} is used here. The Elastic Transform implementation in Albumentations has three parameters: $\alpha$, $\sigma$, affine\_transform. $\sigma$ is fixed at 12 to prevent the synthesis output from excessively curvy, and affine\_transform is fixed at 0.2. Examples of $s$ is shown in the right column of Fig. \ref{dil_samples2} as blue regions.
Finally, $s$ and $p$ are compared by calculating the recall value $r$ (step 10). In order to emulate rushed human annotators, this calculated recall value should neither be too high nor too low. For the annotations generated for the paper, the upper and the lower bounds are chosen to be 0.975 and 0.925, respectively (step 15).
An optimal range of $\alpha$ that achieves the recall target varies among images, so we search for it by first initializing the range to be between 10 to 10000 (steps 3-4), repeatedly drawing from a random distribution (step 8), and narrowing the search space after each trial (steps 11-14).

\begin{algorithm}
    \DontPrintSemicolon
    \SetKwFor{ForAll}{forall}{do}{end forall}%
    \SetKwFor{RepTimes}{repeat}{times}{end}
    \SetKwInOut{Input}{Input} \SetKwInOut{Output}{Output}

    \Input{$P$ \tcp*{Set of Precise Annotations}}
    \InputXX{$n_{dil}$ \tcp*{Number of dilation to apply}}
    \Output{$S$ \tcp*{Set of Synthesized Annotations}}
    $S \gets \emptyset$ \tcp*{Initialize $S$ to be empty}
    \ForAll{$p \in P$}{
        $\alpha_L \gets 10$\;
        $\alpha_U \gets 10000$\;
        \RepTimes{$n_{dil}$}{
            $d \gets Dilate(p)$\;
        }
        \Repeat{$0.925 \leq r \leq 0.975$}{
            Sample random integer $\alpha$ between [$\alpha_L, \alpha_U$]\;
            $s \gets ElasticTransform(\alpha, d)$\;
            $r \gets Recall(s, p)$\tcp*{Calculate recall}
            \If{$r \geq \alpha_U$}{$\alpha_L \gets \alpha$}
            \If{$r \leq \alpha_L$}{$\alpha_U \gets \alpha$}
        }
        $S \gets s$\tcp*{Append to $S$}
    }
    \Return S
    \caption{Annotation Synthesis Pipeline}
    \label{synthesis_algo}
\end{algorithm}

\begin{table*}
	\centering
	\caption{F1-scores for training with the Dilation Annotation. Numbers below model names correspond to $n_{dil}$, the number of dilation operations applied to synthesize the Dilation Annotation. $n_{dil}=0$ corresponds to the Precise Annotation.}

    \newcolumntype{B}{>{\centering\arraybackslash} m{0cm} }
    \newcolumntype{D}{>{\centering\arraybackslash} m{0.35cm}}
    \newcolumntype{C}{>{\centering\arraybackslash} m{0.55cm}}
    \setlength{\tabcolsep}{2pt}
	\begin{tabular}{@{\extracolsep{4pt}}Dc|CCCCCCCCCCCCCCCCCCCC@{}}
		\hline
		 && \multicolumn{5}{c}{Inoue Light} & \multicolumn{5}{c}{Inoue} & \multicolumn{5}{c}{DeepCrack} & \multicolumn{5}{c}{DeepLab V3+} \\
		\cline{3-7} \cline{8-12} \cline{13-17} \cline{18-22}
    	    &MiB&0 &1 &2 &3 &4 &0 &1 &2 &3 &4 &0 &1 &2 &3 &4 &0 &1 &2 &3 &4   \\ \hline
	     \multirow{2}{*}{Aigle} &\checkmark& 0.691&0.788&0.787&0.776&0.768&0.803&0.806&0.800&0.796&0.775&0.787&0.772&0.781&0.789&0.774&0.806&0.794&0.787&0.792&0.765 \\
	                                      && 0.477&0.501&0.434&0.419&0.416&0.725&0.516&0.468&0.407&0.366&0.724&0.579&0.483&0.411&0.323&0.616&0.511&0.465&0.409&0.408 \\ \hline
	     \multirow{2}{*}{CFD}   &\checkmark& 0.623&0.607&0.580&0.567&0.557&0.670&0.615&0.588&0.572&0.558&0.656&0.636&0.601&0.588&0.574&0.666&0.628&0.592&0.575&0.563 \\
	                                      && 0.587&0.599&0.539&0.517&0.498&0.678&0.627&0.583&0.555&0.533&0.657&0.603&0.540&0.501&0.445&0.667&0.635&0.516&0.431&0.372 \\ \hline
	     \multirow{2}{*}{DCD}   &\checkmark& 0.824&0.821&0.813&0.810&0.805&0.822&0.831&0.822&0.817&0.813&0.845&0.831&0.824&0.833&0.823&0.848&0.818&0.825&0.817&0.797 \\
	                                      && 0.782&0.756&0.699&0.698&0.640&0.795&0.783&0.734&0.709&0.674&0.841&0.799&0.776&0.769&0.749&0.842&0.786&0.764&0.685&0.642  \\ \hline
	\end{tabular}
	\label{table_dil_result}
\end{table*}

\begin{table*}
	\centering
	\caption{F1-scores for training with the Rough and Rougher Annotations.}

    \newcolumntype{D}{>{\centering\arraybackslash} m{0.35cm}}
    \newcolumntype{C}{>{\centering\arraybackslash} m{1cm}}
    \setlength{\tabcolsep}{2pt}
	\begin{tabular}{@{\extracolsep{4pt}}Dc|CCCCCCCCCCCC@{}}
		\hline
		 && \multicolumn{3}{c}{Inoue Light} & \multicolumn{3}{c}{Inoue} & \multicolumn{3}{c}{DeepCrack} & \multicolumn{3}{c}{DeepLab V3+} \\
		\cline{3-5} \cline{6-8} \cline{9-11} \cline{12-14}
    	    &MiB&Precise &Rough &Rougher &Precise &Rough &Rougher &Precise &Rough &Rougher &Precise &Rough &Rougher  \\ \hline
	     \multirow{2}{*}{Aigle} &\checkmark& 0.691&0.773&0.781&0.803&0.816&0.802&0.787&0.816&0.808&0.806&0.773&0.811 \\
	                                      && 0.477&0.433&0.383&0.725&0.522&0.435&0.724&0.538&0.410&0.616&0.487&0.407 \\ \hline
	     \multirow{2}{*}{CFD}   &\checkmark& 0.623&0.621&0.591&0.670&0.631&0.597&0.656&0.629&0.590&0.666&0.641&0.599 \\
	                                      && 0.587&0.586&0.519&0.678&0.625&0.588&0.657&0.624&0.536&0.667&0.649&0.506 \\ \hline
	     \multirow{2}{*}{DCD}   &\checkmark& 0.824&0.824&0.811&0.822&0.836&0.813&0.845&0.835&0.830&0.848&0.842&0.823 \\
	                                      && 0.782&0.770&0.701&0.795&0.796&0.722&0.841&0.813&0.794&0.842&0.825&0.710  \\ \hline
	\end{tabular}
	\label{table_man_result}
\end{table*}

\subsection{Results on the Dilation, Rough, Rougher Annotations}

Results on the Dilation, Rough, Rougher Annotations are summarized in Fig. 4 and 5 of the original paper. Although these graphs give intuitive understandings of the overall trend, they lack the precise values of each data point, making comparisons with future research difficult. To resolve this issue, we present the table version of the two figures in Tables \ref{table_dil_result} and \ref{table_man_result}.

\subsection{Test-Train Split for Aigle and CFD}

Although both Aigle and CFD are well-known public datasets used in many crack detection literature, we were not able to find any official test-train split for those datasets. So we record our split in this section to promote future research.

\subsubsection{Aigle Test Data}
C18bor, E17aor, E17bor, F01aor, F02aor, F04bor, F05bor, F08bor, F09aor, F10bor, F12bor, F13aor, F14aor, F16aor.

\subsubsection{CFD Test Data}
002, 004, 005, 006, 014, 016, 018, 024, 025, 027, 028, 029, 033, 036, 037, 038, 041, 044, 047, 049, 053, 059, 060, 062, 064, 066, 073, 074, 076, 077, 078, 085, 090, 091, 093, 094, 096, 098, 102, 104, 108, 110, 111, 112, 114, 116, 118.

\end{document}